\documentclass[conference]{IEEEtran}
\IEEEoverridecommandlockouts
\usepackage{cite}
\usepackage{amsmath,amssymb,amsfonts}
\usepackage{algorithmic}
\usepackage{graphicx}
\usepackage{textcomp}
\usepackage{xcolor}
\usepackage{tabularx}
\usepackage{blindtext}
\usepackage{multirow}
\usepackage{graphicx}
\usepackage{url}

\def\BibTeX{{\rm B\kern-.05em{\sc i\kern-.025em b}\kern-.08em
    T\kern-.1667em\lower.7ex\hbox{E}\kern-.125emX}}
    
\usepackage{fancyhdr}
\begin{document}

\title{Using Clinical Drug Representations for Improving Mortality and Length of Stay Predictions}

\author{\IEEEauthorblockN{Batuhan Bardak and Mehmet Tan\thanks{}}
\IEEEauthorblockA{\textit{Department of Computer Engineering}\\ 
\textit{TOBB University of Economics and Technology}\\
Ankara, Turkey}
Email:\{b.bardak, mtan\}@etu.edu.tr}

\maketitle
\thispagestyle{plain}
 \fancypagestyle{plain}{
 \fancyhf{} 
 \fancyfoot[L]{978-1-6654-0112-8/21/\$31.00~\copyright2021~IEEE} 
 \renewcommand{\headrulewidth}{0pt}
 \renewcommand{\footrulewidth}{0pt}
 }

\begin{abstract}
Drug representations have played an important role in cheminformatics. However, in the healthcare domain, drug representations have been underused relative to the rest of Electronic Health Record (EHR) data, due to the complexity of high dimensional drug representations and the lack of proper pipeline that will allow to convert clinical drugs to their representations. Time-varying vital signs, laboratory measurements, and related time-series signals are commonly used to predict clinical outcomes. In this work, we demonstrated that using clinical drug representations in addition to other clinical features has significant potential to increase the performance of mortality and length of stay (LOS) models. We evaluate the two different drug representation methods (Extended-Connectivity Fingerprint-ECFP and SMILES-Transformer embedding) on clinical outcome predictions. The results have shown that the proposed multimodal approach achieves substantial enhancement on clinical tasks over baseline models. Using clinical drug representations as additional features improve the LOS prediction for Area Under the Receiver Operating Characteristics (AUROC) around \%6 and for Area Under Precision-Recall Curve (AUPRC) by around \%5. Furthermore, for the mortality prediction task, there is an improvement of around \%2 over the time series baseline in terms of AUROC and \%3.5 in terms of AUPRC. The code for the proposed method is available at https://github.com/tanlab/MIMIC-III-Clinical-Drug-Representations.

\end{abstract}


\section{Introduction}
\label{sec:intro}

The recent advancements in deep learning have led to an increase in the number of studies in the healthcare domain, and many researchers utilize novel deep learning and statistical models on a wide range of clinical tasks. The adoption of Electronic Health Records (EHR) presents a rich opportunity for valuable clinical predictive models. In modern EHR systems, for each admitted patient, hospital intensive care units record the data related with the patient including but not limited to laboratory test results, clinical notes, prescriptions, and disease history. 

\renewcommand{\arraystretch}{1.2}
\begin{table*}[!t]
\caption{Summary statistics of the original MIMIC-III dataset, and the final cohort that is used in this study.}
\centering
\resizebox{\textwidth}{!}
{\begin{tabular}{l c c c}

\hline

& \# of Patient & \# of hospital admission & \# of ICU admission \\ \hline
MIMIC-III ( $>$ 15 years old)                                       & 38,597         & 49,785                    & 53,423              \\ 
MIMIC-Extract                                    & 34,472         & 34,472                    & 34,472               \\ 
MIMIC-Extract (at least 24+6 (gap) hours patient) & 23,937         & 23,937                    & 23,937               \\ \hline
\textbf{Final Cohort} (After drug elimination)                        & 22,013         & 22,013                   &  22,013             \\ \hline
\end{tabular}}

\label{table:summary-datasets}
\end{table*}

As a result of the rich context of EHR, tremendous efforts have been made to predict several important clinical tasks such as mortality~\cite{lin2019predicting}, length of stay (LOS)~\cite{turgeman2017insights}, hospital readmission~\cite{mahmoudi2020use}, ICD code assignment~\cite{moons2020comparison}, medical concept extraction~\cite{meystre2008extracting, savova2010mayo} and more. In this paper, we focus on LOS ( $>$3 \& $>$7) and mortality (in-hospital \& in-ICU) prediction tasks. Accurate estimation of patient LOS at ICU provides useful information regarding cost containment and improves the utilization of hospital resources~\cite{garg2012intelligent}. Mortality prediction is the second clinical task that we focus on which is one of the primary concerns of any healthcare system. The probability of mortality can give the measure of the effectiveness of the treatment~\cite{afessa2004identifying}. In addition, it can provide an evaluation of suitability of novel treatments for the patients~\cite{russell2015assessment}, and it can guide doctors to the appropriate treatment for patient benefit. Until the last few years, most of the demonstrated techniques~\cite{song2018attend, suresh2018learning} for training predictive models for clinical data were based on structured data such as time-series signals in EHR data. Recently, there has been increasing interest in utilizing different modalities of data together to improve the performance of clinical models. One of the most important modality is clinical notes which contain valuable information for predicting the patient's health progress. As a result, some studies~\cite{jin2018improving, khadanga2019using, bardak2021improving, bardak2021prediction} leverage from clinical notes in addition to the time-series data in multimodal approach. Instead of using clinical notes as another modality, we focus on prescription information for patients. The prescriptions data in Medical Information Mart for Intensive Care (MIMIC-III)~\cite{johnson2016mimic}, which is the most popular publicly available EHR dataset up to date, consists of medication related order records such as drug name, National Drug Code (NDC), dosage, and more. Drugs given to patients can contain useful information for patient mortality and LOS prediction. Also, in the field of cheminformatics, learning an effective drug representation is a hot topic and these drug representations have been utilized with deep learning methods for several applications~\cite{ozturk2018deepdta, yang2019analyzing}. These findings have motivated us to employ molecular representations of clinical drugs in the proposed method to improve clinical outcomes.


In this study, we propose a multimodal neural network which comprises 1D Convolutional Neural Network (CNN) for clinical drug representations and Gated Recurrent Unit (GRU) for clinical time-series features. The model results show that using clinical drug representations with time-series based features improves the mortality and LOS predictions. We also investigate the effect of using different drug representations such as extended-connectivity fingerprint (ECFP)~\cite{rogers2010extended} and pre-trained Simplified Molecular Input Line Entry System (SMILES)-Transformer~\cite{honda2019smiles}. The key contribution of this work is two folds. First, because of each drug representation methods are constructed to capture different features of the drugs, we use ECFP and pre-trained SMILES Transformer representations in the proposed model. We believe that using these representations for drugs will enable many other studies to use clinical drugs more effectively in healthcare studies. Second, we propose a multimodal neural network that uses clinical drug representations in order to improve the accuracy of mortality and LOS predictions. To our best knowledge, this will be the first study in the healthcare domain that uses clinical drug representations (chemical structure of drugs) with the time-series based EHR data for predicting clinical outcomes. 


    
    

The paper is organized as follows: First, similar studies regarding that the clinical tasks and drug representations are summarized. In Section~\ref{sec:methods}, the dataset and the methods that were used in this study are described. Following that, experimental results are reported in Section~\ref{sec:experiments} and the paper is concluded with a discussion of model results and conclusion.

\begin{figure*}[!t]
  \centering
  \includegraphics[width=\linewidth]{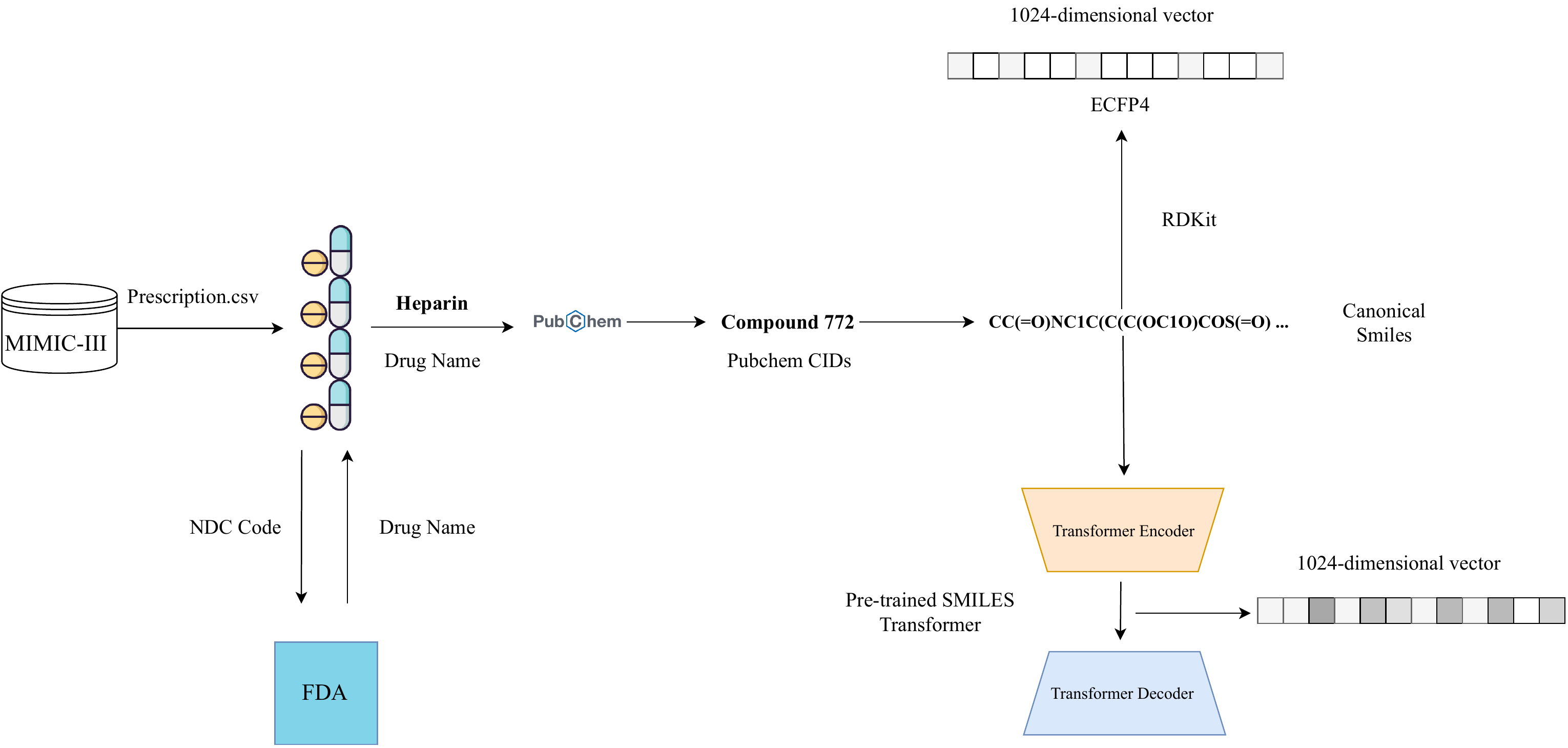}
  \caption{Pipeline for clinical drug extraction from MIMIC-III and converting them into ECFP and pre-trained SMILES Transformer representations.}
  
  \label{fig:drug-pipeline}
\end{figure*}

\section{Related Work}
\label{sec:related-work}

There is an extensive number of research on clinical domain using machine learning algorithms. The ones closely related with this study are explained in this section. The early attempts that use machine learning to EHR data generally focus on using temporal order of events, diagnoses categories for a subsequent visit. One of the earliest works, Doctor AI by Choi et al.~\cite{choi2016doctor} used GRU to predict multi label diagnosis for the next visit by using preceding diagnosis, medication, and procedure codes. REverse Time AttentIoN (RETAIN)~\cite{choi2016retain} is another study by Choi et. al that used attention mechanisms to add interpretability to their predictive models. Two-level neural attention model is used to detect influential past patterns. Another study,~\cite{lipton2015learning} investigated the performance of LSTM model to classify 128 diagnoses by using 13 frequently but irregularly sampled vital signs from the patient in pediatric ICU. 

In this work, we focus on mortality and LOS which are vital clinical prediction tasks for healthcare. Early works for mortality prediction showed that simple machine learning algorithms with hand-crafted features obtain accurate results~\cite{kim2011comparison, dybowski1996prediction, celi2012database}. Also, there are several severity scoring system methods used to determine the severity of illness such as chronic health evaluation (APACHE)~\cite{knaus1981apache}, simplified acute physiology score (SAPS-II)~\cite{le1993new}, and the sequential organ failure assessment (SOFA)`\cite{vincent1996sofa}. Mortality prediction can be made by using these calculated severity scores. On the other hand, as in many fields, deep learning architectures have also become popular for mortality prediction. In~\cite{awad2017early}, they used ensemble learning technique with demographics, vital signs, and laboratory test data to make early mortality prediction. More recently, convolutional neural network and gradient boosted algorithms have been used by Darabi et al.~\cite{darabi2018forecasting} to predict 30 days mortality risk. 
Sotoodeh et. al.~\cite{sotoodeh2019improving} used a hidden Markov model framework that utilizes the first 48 hours of physiological measurements to predict LOS. A comprehensive survey on mortality and LOS prediction in ICU can be found in~\cite{awad2017patient}.

One of the common challenges in healthcare research today is preprocessing EHR data. There are few studies~\cite{harutyunyan2019multitask, wang2020mimic, purushotham2018benchmarking} that proposed standardized preprocessing steps such as unit conversion, outlier detection, feature engineering, and data assembly for MIMIC-III. Especially, the most recent effort in this area, MIMIC-Extract~\cite{wang2020mimic} proposed an open source pipeline for transforming MIMIC-III data and public benchmarks for different clinical tasks. In this study, MIMIC-Extract pipeline is used to featurize MIMIC-III data. 

Recently, multimodal learning approach is heavily used in healthcare domain which employes multiple data sources to predict clinical tasks. The studies have shown that it is possible to improve clinical outcome predictions by using multimodal methods. Khadanga et al.~\cite{khadanga2019using} showed that combining clinical notes as another modality improves the performance of three clinical tasks (in-hospital mortality, decompensation, LOS). In another work~\cite{jin2018improving}, name entity recognition (NER) algorithm is applied to clinical notes to extract medical entities which are used together with time-series data. In this work, instead of using clinical notes as an additional feature, clinical drug representations are used in multimodal approach.

Drug representations have been used for several cheminformatic applications such as drug discovery~\cite{chan2019advancing}, drug-drug interaction~\cite{ryu2018deep}, and compound-protein affinity~\cite{ozturk2018deepdta} prediction. To use drug representations in machine learning models, molecular data should be transformed to a suitable format. Learning expressive representation of molecule structure is a fundamental problem due to the complex nature of it. To address this issue, researchers proposed various methods based such as based on hand-crafted fingerprints~\cite{xue2000molecular} and graph neural network~\cite{gilmer2017neural}. Over the past three years, transformer-based architectures~\cite{devlin2018bert, vaswani2017attention} in Natural Language Process (NLP) give a state-of-the-art performance on several different NLP tasks. This pre-trained approach is also employed for learning molecular representations~\cite{chithrananda2020chemberta, wang2019smiles}. In this study, the experiments are conducted with two different drug representation methods. First, pre-trained SMILES-Transformer model is used which consists of an encoder-decoder network with 4 Tranformer blocks. The second representation that used in the experiments is ECFP which is designed for molecular characterization and structure-activity modeling.

\section{Methods}
\label{sec:methods}

This section begins with the description of dataset and feature transformation process. Then, the details of time-series baseline model used in this work are given. Finally, the proposed multimodal deep learning architecture which uses drug representations in addition to clinical time-series features is introduced.

\subsection{Data}
\label{subsubsec:data}

The data set used in this research is extracted from publicly available MIMIC-III~\cite{johnson2016mimic}. There are 46,520 patients and 61,532 ICU admissions in this database. We use a novel open source MIMIC-Extract~\cite{wang2020mimic} pipeline to extract and preprocess the data from MIMIC-III. MIMIC-Extract pipeline makes unit conversion, outlier detection, feature aggregation, and some other preprocessing steps into MIMIC-III and constructs a rich cohort of 34,472 patients and 104 time-series features. The 104 features that are extracted with MIMIC-Extract contains clinically aggregated time-series variables related with both vitals signs such as heart rate, blood pressure and laboratory test results such as white blood cell counts, fraction inspired oxygen and there is no any prescription drugs are included.

In addition to the output features of the MIMIC-Extract pipeline, to improve the clinical task predictions, we extract drug information from prescription data in MIMIC-III and utilize them in the proposed multimodal approach. The details of drug information extraction and usage process are explained in Section~\ref{subsubsec:drug-rep}. It should be noted that only the first 24 hours of patient's data after ICU admission is considered. We exclude patients from the final cohort who had not taken any drug in the first 24 hours after ICU admission. We split the data with 70\% 10\% 20\% ratio (train/val/test) and use stratified splitting technique for all tasks. All statistics of the datasets are given in Table~\ref{table:summary-datasets}.

\begin{figure*}[!t]
  \centering
  \includegraphics[width=\linewidth]{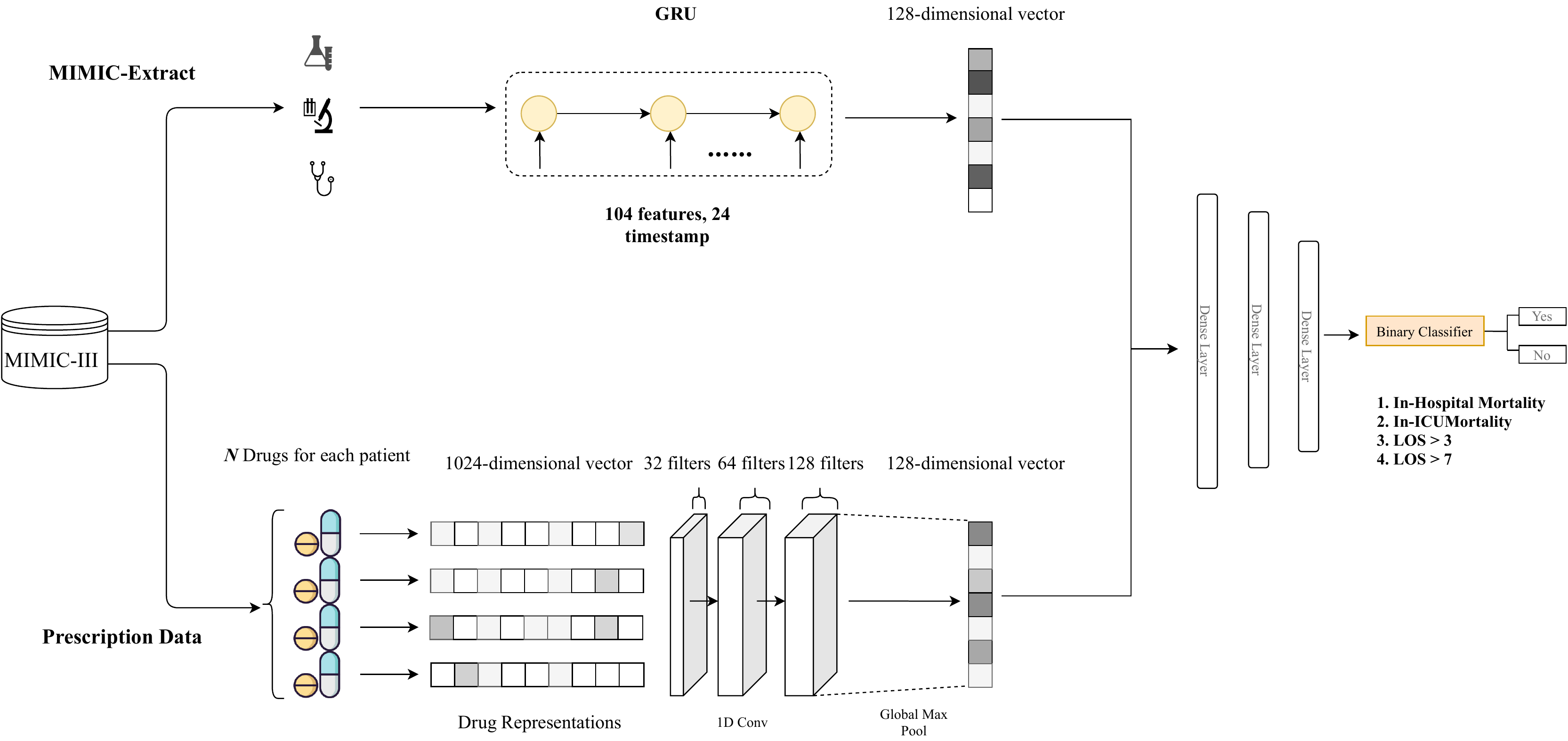}
  \caption{Overview of Proposed multimodel architecture for predicting the In-Hospital Mortality, In-ICU Mortality, LOS $>$ 3, and LOS $>$ 7. To extract timeseries features, we use MIMIC-EXTRACT pipeline and fed these features through GRU. 1D CNN is applied to drug representations. In the final layer, we concatenate features that extracted from timeseries and drug representations and fed through fully connected layers to predict 4 different binary clinical tasks.}
  \label{fig:proposed-model}
  
\end{figure*}

\subsubsection{Prescription Data}
\label{subsubsec:drug-rep}

The prescription data in MIMIC-III provides information about prescribed medications. This information contains the name of the drug, coded identifiers like Generic Sequence Number (GSN), and National Drug Code (NDC), the drug strength, dose information, and the route of administration. There are also start and end dates which specify the date period of a drug for each patient however the quality of this feature is unclear so we omit it.
The example prescription data is shown in Table~\ref{tab:ex-pres}. In MIMIC-III, there are 4525 unique drug names, 4504 NDC codes, and 2863 unique generic drug names. 

\setlength{\tabcolsep}{1.6pt}
\renewcommand{\arraystretch}{1.8}
\begin{table}[!h]
\caption{Example drug, generic drug name and NDC codes in prescription data.}
\centering
\begin{tabular}{c c c}
\hline
\textbf{Drug Name} &\textbf{Generic Drug Name} & \textbf{NDC Code}  \\ \hline
Heparin & Heparin Sodium & 63323026201.0 \\  \hline
Acetaminophen &  Acetaminophen & 182844789.0\\  \hline
Lorazepam  & Lorazepam & 594091985307.0 \\  \hline
Morphine Sulfate  & Morphine Sulfate (Syringe) & 409176230.0\\  \hline
\end{tabular}

\label{tab:ex-pres}
\end{table}

\vspace{2mm}

\noindent\textbf{Clinical Drug to SMILES Pipeline.} We introduce a pipeline in Figure~\ref{fig:drug-pipeline} to convert clinical drug names and NDC codes to drug representations. In the first part of the pipeline, we extract the unique generic drug names, unique drug names, and NDC codes which are related with the patients in the final cohort. In the final cohort, there exist 2255 unique drug names, 3273 NDC codes, and 1688 unique generic drug names. Then, we search for general drug names in Pubchem~\cite{bolton2008pubchem} compound database to find their Pubchem CIDs using open source pubchempy\footnote{\url{https://pubchempy.readthedocs.io/en/latest/}} library. The fact that some of the generic drug names are written informally makes it difficult to find them in Pubchem database. In such cases, we try to use drug names in prescription data. If this method does not work either, lastly we use NDC codes and search them from the Food and Drug Administration (FDA) database,\footnote{\url{https://open.fda.gov/data/ndc/}}. The corresponding Pubchem CIDs are used to extract text based representation system for molecules which is named SMILES~\cite{weininger1988smiles}. In order to utilize these SMILES strings as an input to the deep learning models, we must transform them into a suitable vector representation. In the literature, several approaches utilize the discrete molecular representation, such as SMILES strings, and convert it into low-dimensional continuous representations. In this work, two different representation methods are used, ECFP and pre-trained SMILES-Transformer model. These drug representations are combined with time-series clinical features to improve the success of clinical outcomes.  


\noindent\textbf{Representations of Drugs in Latent Space.} Many applications in cheminformatics field, molecular descriptors, and fingerprints are encoded to low dimensional vectors. Learning an effective drug representation in molecular space is extremely challenging, due to the large search space, and the complex, unstructured property of drugs. There are lots of proposed approaches such as convolutional neural networks on graphs~\cite{duvenaud2015convolutional}, Coulomb matrices~\cite{rupp2012fast} to transform molecular representation into effective continuous embedding vectors. In this study, we choose to use a common text based SMILES representation for clinical drugs and transform it into two different vector representations. First, we employ the open source cheminformatics library RDKit~\footnote{\url{https://www.rdkit.org/}} to transform SMILES string into ECFP which is a popular molecular fingerprints. To create a fixed-length ($1024$ for this study) binary vector where 1 indicates the existence of the assigned substructure and 0 for the absence, ECFP applies a fixed hash function to the concatenated features of the neighborhoods. Secondly, we use pre-trained SMILES Transformer~\cite{honda2019smiles} model to convert clinical drug names into the embedding vector. 861,000 unlabeled SMILES are randomly sampled from ChEMBL24 dataset and used for unsupervised training. SMILES Transformer compromises of encoder-decoder network with 4 Transformer blocks. We extract 1024-dimensional fingerprints from the pre-trained SMILES Transfomer model for each clinical drug SMILES.

\subsection{Time Series Model}

The baseline models are trained using only time-series features which are extracted from the MIMIC-Extract pipeline. We employ both Long Short Term Memory (LSTM)~\cite{hochreiter1997long} and Gated Recurrent Units (GRU)~\cite{chung2014empirical} networks which are types of recurrent neural networks to capture the temporal information between the patient features. Based on time-series model experiments, GRU has shown a slightly better performance than LSTM. In general, GRU cell has two gates, a reset gate $r$ and an update gate $z$. With this novel gate structure, GRU can handle the vanishing gradient problem. Due to these reasons, we use GRU architecture both in baselines and proposed multimodal architecture. We formulate a time-series model as follows. At each step, the GRU takes the current input $x_t$ with its previous hidden state $h_{t-1}$ to generate the current hidden state $h_t$, for $t=1$ to $t=24$. For all four clinical tasks, we use the same GRU architecture and we can write the mathematical formulation as follows:
\begin{equation}
\begin{split}
z_t  = \sigma(W_z x_t + U_z h_{t-1} + b_z) \\
r_t  = \sigma(W_r x_t + U_r h_{t-1} + b_r) \\
\hat{h}_t = \tanh(w_h x_t + r_t \circ U_h h_{i-t} + b_h) \\
h_t = z_t \circ h_{t-1} + (1 - z_t) \circ \hat{h}_t \\
\hat{prediction} = \text{sigmoid}(W_h h_{t}+b_h) \\
\end{split}
\end{equation}
where $z_t$ and $r_t$ respectively represent the update gate and the reset gate, $\hat{h}_t$ the candidate activation unit, $h_t$ the current activation, and $\circ$ represents element-wise multiplication. For predicting the mortality and LOS, we use sigmoid activation function on the top of the one layer GRU with $128$ hidden units.

\renewcommand{\arraystretch}{1.2}
\setlength{\tabcolsep}{8pt}
\begin{table*}[!t]
\caption{Average scores with three different metrics for all four tasks. Proposed model performance comparison with best baseline model. We select the highest score for each metric and each clinical task from all methods.}
\centering
\resizebox{\textwidth}{!}{\begin{tabular}{l c c c c c}
\hline
Task & Features & Drug Embedding & AUROC & AUPRC & F1\\ \hline
\multirow{2}{*}{\textbf{In-Hospital Mortality}} 

& Time-series Baseline & - & 85.69 $\pm$ 0.007 & 47.32 $\pm$ 0.014 & 43.05 $\pm$ 0.011\\
\cline{2-6}
&                  & Smiles-Transformer & 86.97 $\pm$ 0.004 & 50.09 $\pm$ 0.007 & 43.84 $\pm$ 0.007\\
 & Proposed Model        &  ECFP & \textbf{87.66} $\pm$ 0.004 & \textbf{51.37} $\pm$ 0.006 & \textbf{44.90} $\pm$ 0.008\\
 
\hline
\hline

\multirow{2}{*}{\textbf{In-ICU Mortality}} 

& Time-series Baseline & - & 86.73 $\pm$ 0.006 & 42.33 $\pm$ 0.012 & 42.01 $\pm$ 0.011\\
\cline{2-6}
 &                  & Smiles-Transformer & 88.33 $\pm$ 0.001 & 46.14 $\pm$ 0.007 & 45.28 $\pm$ 0.008\\
  & Proposed Model             &  ECFP & \textbf{88.98} $\pm$ 0.005 & \textbf{46.73} $\pm$ 0.007 & \textbf{45.68} $\pm$ 0.014\\
   
\hline
\hline

\multirow{2}{*}{\textbf{LOS $>$ 3 Days}} 

& Time-series Baseline & - & 68.96 $\pm$ 0.002 & 62.76 $\pm$ 0.003 & 57.09 $\pm$ 0.007\\
\cline{2-6}
 &                  &  Smiles-Transformer & 69.70 $\pm$ 0.004 & 63.19 $\pm$ 0.005 & 57.60 $\pm$ 0.006\\
  & Proposed Model         &  ECFP & \textbf{71.70} $\pm$ 0.004 & \textbf{65.46} $\pm$ 0.004 & \textbf{59.79} $\pm$ 0.005\\
\hline
\hline

\multirow{2}{*}{\textbf{LOS $>$ 7 Days}} 

& Time-series Baseline & - & 67.39 $\pm$ 0.022 & 15.09 $\pm$ 0.017 
& 20.40 $\pm$ 0.024\\
\cline{2-6}
 &                  &  Smiles-Transformer &73.32 $\pm$ 0.008 & \textbf{20.33} $\pm$ 0.007 & 26.48 $\pm$ 0.005 \\
  & Proposed Model             &  ECFP & \textbf{73.54} $\pm$ 0.008 & 
  19.75 $\pm$ 0.012 & \textbf{27.55} $\pm$ 0.006\\
\hline

\end{tabular}}

\label{tab:proposed-results}
\end{table*}

\subsection{Proposed Approach}

In this work, all clinical problems are formulated as a binary classification problem and the main objective is effectively using clinical drug representations to improve clinical outcome predictions. To make a fair comparison with the baseline models, we use the same GRU architecture for clinical time-series features and use 128 layer GRU architecture to extract feature map. Convolutional Neural Networks (CNN) models have capability to capture local spatial features that are used in different kind of problems in the literature~\cite{lecun1998gradient}. CNN is also to be effective for extracting feature map on grid-like data, such as images and texts
. We try to adopt a convolutional approach similar to~\cite{kim2014convolutional} and take an advantage of 1D convolutional layers as a feature extractor on clinical drug representations. Let $d_i \in \mathbb{R}^{k}$ be the k-dimensional ($1024$ in our experiments) drug representation corresponding to the $i$-th drug in the drug list for each patient. The number of drugs taken by each patient is varying, however, we padded where it necessary to make fixed-length vectors where it denotes with $n$. These drug representations $d_i \in \mathbb{R}^{k}$ are combined vertically and each patient is represented by a matrix $M \in {R}^{k*n}$ where rows are filled with clinical drug representations. To sum up, the patient clinical drug matrix is represented as:

\begin{equation}
\mathbf{d}_{1:n} = \mathbf{d}_1 \otimes \mathbf{d}_2 \otimes \ldots \otimes \mathbf{d}_n
\end{equation}

where $\otimes$ is the concatenation operator and $d$ refers to the representation of the drug and $n$ is the number of drug for each patient which is selected $64$ in the experiments. Feature extraction procedure on drugs is similar to the one in~\cite{ozturk2018deepdta}. We consecutively stack three 1D convolutional layers with filter size $32$, $64$, $128$. The kernel size is selected as $3$ for all three 1D-convolutional layers. At the end of the last convolutional layer, to capture the most important features, we apply a max-pooling layer over the feature map. The extracted features from the max-pooling layer are concatenated with the features from one layer GRU with 128 hidden units and fed through fully connected layers. We use three fully connected layers in the last part of architecture with $1024$, $512$, and $256$ nodes. The proposed model that combines the clinical drug representations and clinical time-series features is shown in Figure~\ref{fig:proposed-model}.

\section{Experiments}
\label{sec:experiments}

In this part, the evaluation and the implementation details of the models are discussed, and the results of the baseline and the proposed multimodal experiments are introduced.

\subsection{Experimental Setup}
\label{subsec:experimental-setup}


We work on four binary clinical tasks, corresponding to mortality over the ICU stay (class imbalanced) and hospital admission (class imbalanced) and ICU length of stay greater than three (class balanced) and seven days (class imbalanced). All of the models are evaluated with the same three metrics commonly used in imbalanced classification problems: Area Under the Receiver Operating Characteristics (AUROC), Area Under Precision-Recall (AUPRC), and F1 score. The F1 score, a harmonic combination of both $precision$ and $recall$, is defined as follows:

\vspace*{-0.1cm}
\begin{equation}
    F1 = \frac{2*Precision*Recall}{Precision+Recall}
\end{equation}
\vspace*{-0.4cm}

\noindent where $precision$ is a measure of exactness and $recall$ is a measure of completeness. AUROC calculates the overall ranking performance of a classifier and AUPRC is the final metric which is recommended for highly skewed domains~\cite{davis2006relationship}. 
To deal with the data imbalance problem, the class weights are changed during training phase. We give higher weight to the minority class and lower weight to the majority class. For in-hospital mortality, LOS $>3$, and LOS $>7$ tasks, class weight is used as balanced which adjust weights inversely proportional to class frequencies. On the other hand, in-ICU mortality task, the 1:5 ratio is used for class weights. These weight ratios are selected based on empirical experiments.

For all tasks, we use the same cohort and this cohort contains the patient's first 24 hours of ICU data. In all architectures, we use ReLU activation function for adding nonlinearity to models except the final layer. Batch size is selected as 32 and $L_2$ norm for sparsity regularization is selected with the 0.05 scale factor. Adam is used as an optimizer with learning rate $10^{-3}$, and decay rate with $10^{-2}$. Epoch size is set 100 for all models and early stopping is used to prevent overfitting. We also use 0.3 dropout rate at the end of the fully connected layers. All these parameter values are determined by making hyperparameter optimization. We train each model 10 times and report the average performance.


Deep learning algorithms are implemented using Keras, which runs Tensorflow on its backend. All experiments were performed on a computer with NVIDIA Tesla K80 GPU with 24GB of VRAM, 378 GB of RAM, and Intel Xeon E5 2683 processor. The full code of this work is available at https://github.com/tanlab/MIMIC-III-Clinical-Drug-Representations.

\subsection{Experimental Results}
\label{subsec:experimental-results}

To show the effectiveness of using drug representations on clinical outcomes, we first work with only time-series features which are extracted from MIMIC-Extract pipeline. These extracted time-series features are modeled using a GRU architecture. The proposed multimodal results against the best scores taken from time-series GRU models for all four tasks are compared. As seen from results in Table~\ref{tab:proposed-results}, GRU model achieves high AUROC, AUPRC, and F1 scores for many different settings. In response to this, using drug representation information as an additional feature in multimodal setting improves all clinical task results. For the in-hospital mortality prediction task, there is an improvement of~\%2 AUROC,~\%4 AUPRC, and \%2 for F1 score. We see~\%2 AUROC,~\%4.5 AUPRC, and~\%3.5 F1 score improvement for in-ICU mortality. For LOS tasks, the improvement is even better than mortality tasks. In LOS $>3$ task, we observe~\%2 AUROC,~\%3 AUPRC, and~\%2.5 F1 score improvement. Lastly, the use of drug representations makes the greatest improvement in the LOS $>$ 7 clinical task. For all metrics, the model scores are increased by approximately~\%6. The salient difference in model performance is a clear indicator of the proposed multimodal has better predictive
capability.

The efficiency of the proposed model is mostly related with the usage of the drug representations and 1D convolutional based multimodal network. Since the CNNs have ability to learn locally connected features, we take the advantage of convolutional layers to learn representations of drugs to capture hidden patterns in the dataset. According to the results of the experiments, using convolution to extract features on drug representations consistently improves the prediction results. To ensure the reliability of all baseline and multimodel networks, the experiments are repeated 10 times with different initialization and the mean performance scores are reported. 

\section{Discussion}
\label{sec:discussion}

We demonstrated a method to leverage clinical drug representations to improve four binary clinical prediction tasks on MIMIC-III data. The model results illustrated in Table~\ref{tab:proposed-results} show the benefit of using drug representations as an additional feature. The proposed multimodal methods that use drug representations with time-series signals consistently outperform all baselines and achieve better performance. While numerous studies in the literature address various approaches to predict mortality and LOS, as our knowledge, this is the first work that uses the molecular representations of clinical drugs to predict clinical outcomes.

The results of the experiments provide evidence for the promising performance of multimodal approach compare to time-series baseline models. Further, the experiments can give us an opportunity to compare the performance of different methods for molecular drug representations. As shown in Table~\ref{tab:proposed-results}, nearly for all tasks, ECFP representations give better performance than pre-trained SMILES-Transformer. Apart from the fact that, ECFP is a successful method, SMILES-Transformer is trained with a dataset which has a different distribution than clinical drugs that may lead to a performance decrease. In the future, a variety of different drug representation methods that are currently presented in the literature can be tested in a similar approach. We believe that using drug information in this manner can also pave the way for new approaches that assist clinical outcome predictions.

There are many attempts in the literature to predict mortality and LOS. Harutyunyan et al.~\cite{harutyunyan2019multitask} propose an EHR preprocessing pipeline and work on mortality, LOS, decompensation, and phenotyping. They formulate the LOS prediction task as multi-class by dividing the remaining LOS into 10 buckets/classes. They use 17 time-varying clinical aggregate features in their models. Based on their results, LSTM shows better performance than other models. In another study, Khadanga et al.~\cite{khadanga2019using} propose a multimodal network with merging clinical time series embedding and clinical notes embedding together to predict in-hospital mortality, decompensation, and LOS. For time-series features, they use the identical setup with that in \cite{harutyunyan2019multitask}. They reported a significant improvement of their proposed multimodel network over the baseline models. In our study, we apply the MIMIC-Extract pipeline to extract clinical aggregate features and work with 104 features. In addition to this, we utilize the prescription data in MIMIC-III and use a molecular representation of clinical drugs for the first time. 

We completed the last experiment solely using the clinical drug representations without time-series features. This approach results in a poor performance (around less than \%15 - \%30 for different metrics and tasks) on tasks which shows us using drugs only without rich time-series signals are not sufficient to make successful predictions. Due to the poor results, we do not report them in this paper.

\section{Conclusion and Future Work}
\label{sec:conclusion}

This paper investigates mortality and LOS prediction which have been important clinical tasks. Making predictions accurately for these clinical tasks have to potential in improving the quality of patient care and better resource utilization at ICU. In this paper, we use MIMIC-III for EHR dataset, and open source MIMIC-Extract pipeline for extracting clinical time-series features. We propose a pipeline to extract and transform drug information into useful vector representations to be used in deep learning models. This study shows that utilizing clinical drug representations along with clinical time-series features have shown significant improvement for the prediction of mortality and LOS. Throughout several experiments with different drug representations over the patients, ICU data showed that the proposed multimodal neural network outperforms all the baselines in every clinical problem for every evaluation metric (except F1 score on $LOS>7$).

While remarkably accurate results have been achieved, the proposed multimodal method reveals some limitations. First, the Pubchem ID of every drug name in MIMIC-III can not be found conveniently. The reason for this is usually the missing data of some drugs or the informal spelling of these drug names. Second, while the proposed model shows high performance, the interpretability of the model results is low. Third, the drug representations used in this study are not trained specific to clinical drugs.

Several directions can be followed to extend this work. First, we can improve the drug names to SMILES pipeline to find more drug representation. Second, more advanced deep learning architecture with an attention layer will be applied to increase the performance and the interpretability of the model. Understanding which part of the drug representation is targeted by the proposed model when making clinical predictions is vital. This outcome can be analyzed with clinical experts and can pave the way for more advanced studies. Third, the clinical drug representations can be trained with clinical domain specific drugs to improve both the representation quality and the clinical task prediction performance.

\ifCLASSOPTIONcompsoc
  \section*{Funding}
\else
  \section*{Funding}
\fi
This study has been partially funded by The Scientific and Technological Research Council of Turkey (TUBITAK), Grant Number:120E173.


\bibliographystyle{IEEEtran}
\bibliography{references}

\end{document}